\documentclass[conference]{IEEEtran}
\IEEEoverridecommandlockouts

\usepackage{cite}
\usepackage{amsmath,amssymb,amsfonts}
\usepackage{algorithmic}
\usepackage{graphicx}
\usepackage{textcomp}
\usepackage{xcolor}
\usepackage{multirow}
\usepackage[ruled,linesnumbered]{algorithm2e}

\ifCLASSOPTIONcompsoc
\usepackage[caption=false, font=normalsize, labelfont=sf, textfont=sf]{subfig}
\else
\usepackage[caption=false, font=footnotesize]{subfig}
\fi

\makeatletter
\newcommand{\removelatexerror}{\let\@latex@error\@gobble}
\makeatother

\def\BibTeX{{\rm B\kern-.05em{\sc i\kern-.025em b}\kern-.08em
    T\kern-.1667em\lower.7ex\hbox{E}\kern-.125emX}}
\begin{document}

\title{Temporal Calibrated Regularization for Robust Noisy Label Learning}

\author{Dongxian Wu\textsuperscript{1,3} \ \ Yisen Wang\textsuperscript{2} \ \ Zhuobin Zheng\textsuperscript{1} \ \ Shu-Tao Xia\textsuperscript{1,3 \dag}  \\
\textsuperscript{1}Tsinghua University \ \ \ \ \ \ \ \textsuperscript{2}Shanghai Jiao Tong University \ \ \ \ \\
\textsuperscript{3}PCL Research Center of Networks and Communications, Peng Cheng Laboratory
}

\maketitle

\renewcommand{\thefootnote}{\fnsymbol{footnote}}
\footnotetext[2]{Corresponding author: Shu-Tao Xia (xiast@sz.tsinghua.edu.cn)}

\begin{abstract}
Deep neural networks (DNNs) exhibit great success on many tasks with the help of large-scale well annotated datasets. However, labeling large-scale data can be very costly and error-prone so that it is difficult to guarantee the annotation quality (\textit{i.e.}, having noisy labels). Training on these noisy labeled datasets may adversely deteriorate their generalization performance. Existing methods either rely on complex training stage division or bring too much computation for marginal performance improvement. In this paper, we propose a \textbf{T}emporal \textbf{C}alibrated \textbf{R}egularization (TCR), in which we utilize the original labels and the predictions in the previous epoch together to make DNN inherit the simple pattern it has learned with little overhead. We conduct extensive experiments on various neural network architectures and datasets, and find that it consistently enhances the robustness of DNNs to label noise.
\end{abstract}

\section{Introduction}
Deep Neural Networks (DNNs) have demonstrated extraordinary
performance in solving many complex problems, including computer vision \cite{krizhevsky2012imagenet}, natural language processing \cite{vaswani2017attention} and speech recognition \cite{wang2017residual}. The success of DNNs heavily relies on the large-scale well-annotated datasets. However, well-annotated datasets are not always available \cite{torralba200880,tanno2019learning,wu2018multi}. Instead, datasets that contain incorrectly-annotated labels (\textit{i.e.}, noisy labels) are very common. Unfortunately,  DNNs are susceptible to such noisy labels, easily raising the problem of overfitting \cite{zhang2016understanding}. Training accurate CNNs against noisy labels is therefore of great practical importance. 

\begin{figure}
    \centering
        \subfloat[ResNet-32]{
        \includegraphics[width=0.48\linewidth]{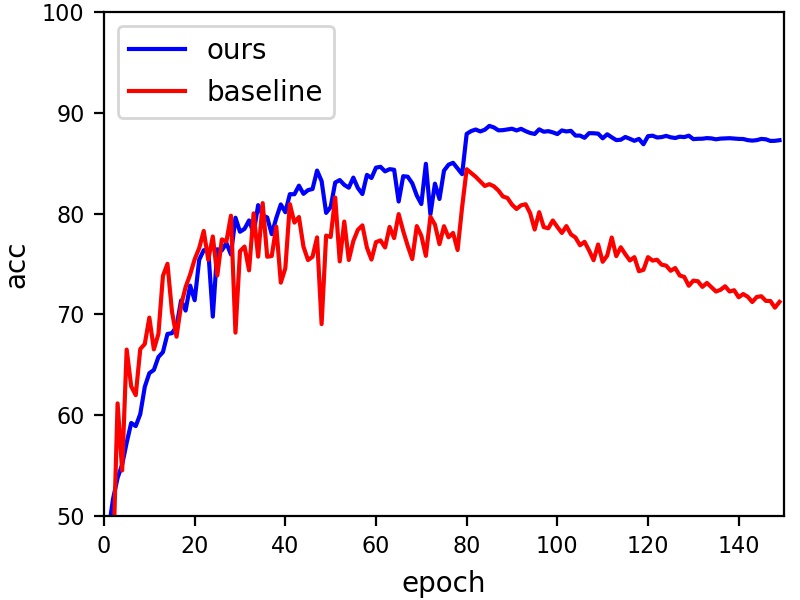}}
        \hfill
	    \subfloat[PreAct-ResNet-18]{
        \includegraphics[width=0.48\linewidth]{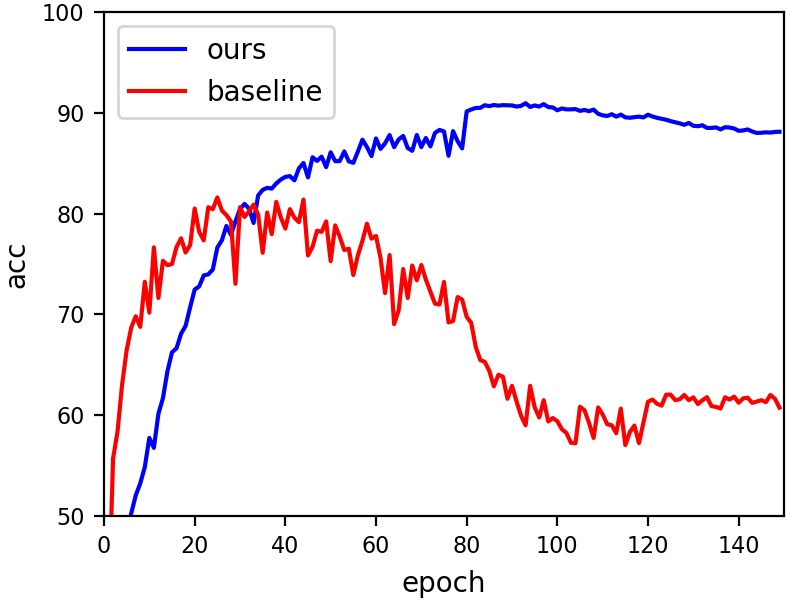}}
    \caption{We trained different architectures in CIFAR-10 with 40\% uniform noise. Although DNN has two stages in learning, the division time relies on its specific setting, which makes two stages method hard to generalize.}
    \label{fig:learning_curve}
\end{figure}

Recent studies have tried to tackle this issue by dividing the DNN training procedure into two stages, \textit{i.e.}, early stage pattern learning and later stage label memorization, either by manual setting \cite{tanaka2018joint} or a complex scheduler \cite{ma2018dimensionality,laine2016temporal}. However, DNNs may start overfiting at any time. For example, as shown in Fig. \ref{fig:learning_curve}, a PreAct-ResNet-18 \cite{he2016identity} starts overfitting around the $30$-\textit{th} epoch (150 epochs in total), while a ResNet-32 \cite{he2016deep} starts overfitting after $80$-\textit{th} epoch. Such methods are limited in practice, since it is hard to predict which epoch start overfitting.
Gradient correction \cite{li2018learning,jenni2018deep} is another kind of method. 
However, it is directly based on the given labels, even when there are noisy labels, thus updating parameters with these noisy gradients will make DNNs easier to overfit. Therefore, there are a certain of work modifying the gradients by the aid of meta-learning \cite{ren2018learning,li2018learning} or bilevel optimization \cite{jenni2018deep}, which require not only extra clean or synthetic noisy samples, but also more forward and backward passes. 

\begin{figure}[t]
\centering
    \includegraphics[width=0.7\columnwidth]{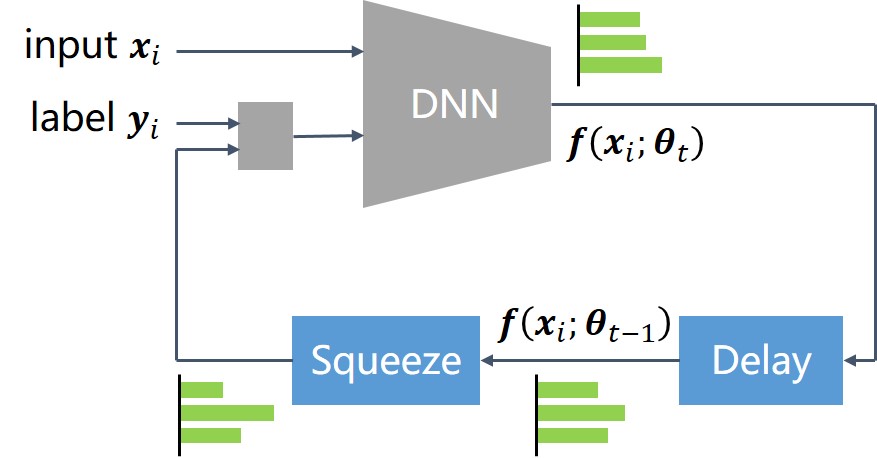}
    \caption{The framework of the proposed Temporal Calibrated Regularization, which consists of a temporal delay (for the reflection loss) and a squeeze function. The network is optimized using the combination of original labels and predictions processed by TCR.}
    \label{fig:diagram_of_our_method}
\end{figure}

In this paper, we propose an extremely simple but effective label correction method, called Temporal Calibrated Regularization (TCR), as shown in Fig. \ref{fig:diagram_of_our_method}. Specifically, we use a convex combination of raw labels and the DNN predictions of the previous epoch as the pseudo-labels for training targets. Besides, we propose to utilize a ``Squeeze'' function, which encourages DNN to have a little higher confidence in its predictions. In TCR, the prediction in the previous epoch is the target for DNN to inherit the simple pattern learned by itself. In this case, DNN can adapt fast without manual stage division or complex scheduler design as we only use the recent historical information. Our method is therefore easy to implement and extend to different situations. 

Our main contributions are:
\begin{itemize}
\item We propose a simple but effective method, Temporal Calibrated Regularization, for robust learning, which considers both the original raw labels and DNN historical predictions so as to inherit the simple pattern learned before and hinder overfitting to noisy labels.
\item To counteract the adverse effects of resistance degradation, we introduce ``Squeeze'' technique to tackle this problem by amplifying the largest prediction to encourage the model to have both distinguishable and high confidence in predicting labels.
\item We conduct extensive experiments on various network architectures and datasets, and demonstrate the competitive performance of the proposed method in image classification tasks compared to state-of-the-art methods..
\end{itemize}

\section{Related Work}

Learning with noisy labels has been widely investigated in \cite{frenay2014classification}, and a variety of approaches have been proposed to robustly train DNNs on noisy datasets, such as regularization and correction. Since \cite{arpit2017closer} showed that dropout can hinder overfitting, other effective regularizations, e.g. mixup \cite{zhang2017mixup} and Bilevel Optimization \cite{jenni2018deep}, were developed. Another approach is to use label correction to eliminate the influence of noisy labels. Reference \cite{reed2014training} replaced the target labels with a convex combination of original labels and current predictions from DNN. Further, A complex combination strategy was utilized based on local intrinsic dimensionality in \cite{ma2018dimensionality}. Joint optimization was proposed in \cite{tanaka2018joint} to jointly optimize network parameters and data labels during training. Moreover, there are approaches to formulate noise models explicitly or implicitly through DNNs \cite{patrini2017making,xiao2015learning,sukhbaatar2014training,goldberger2016training,lv2020matrix}, conditional random field \cite{vahdat2017toward} and knowledge graphs \cite{li2017learning,lu2016big,zhong2019graph}. These noise models are then applied to correct the loss, infer the true labels or assign smaller weights to noisy samples. Other studies introduced robust losses \cite{manwani2013noise,ghosh2015making,van2015learning,natarajan2013learning,zhang2018generalized,wang2019symmetric,ma2020normalized} through noise modeling. However, these methods rely on the modeling assumptions and are limited to generalize to complex situations. They may also request extra clean data or use expensive estimation methods.

Our proposed Temporal Calibrated Regularization (TCR) aims at keeping the simple pattern that DNN learned before and impede ``brutally'' memorizing noisy labels through a label correction method. It also can be regarded as a simplified modifying gradient method.

%-------------------------------------------------------------------------
\section{Temporal Calibrated Regularization}
In this section, we first provide the preliminaries and notations, then followed by the definition of the proposed reflection loss, gradient analysis and resistance degradation in our proposed Temporal Calibrated Regularization (TCR).

\subsection{Preliminaries and Notations}
Column vectors and matrices are denoted in bold (e.g. $\boldsymbol{x}$). $\boldsymbol{1}$ is a vector of all ones and $\boldsymbol{e}^i$ denoted the $i$th standard canonical vector in $\mathbb{R}^c$, i.e. $\boldsymbol{e}^i \in \{ 0, 1 \}^c, \boldsymbol{1}^\top \boldsymbol{e}^i = 1$. 
In a typical $c$-class classification problem, we are given a set of $n$ training inputs $\{\boldsymbol{x}_1, \boldsymbol{x}_2, \cdots, \boldsymbol{x}_n \}$ with their corresponding labels $\{ \boldsymbol{y}_1, \boldsymbol{y}_2, \cdots, \boldsymbol{y}_n \}$ within label space $\mathcal{Y} = \{\boldsymbol{y}: \boldsymbol{y} \in [0, 1]^c, \boldsymbol{1}^\top \boldsymbol{y}=1\}$. Given an input $\boldsymbol{x}_i$, the output vector of the DNN with parameter $\boldsymbol{\theta}$ is $\boldsymbol{h}(\boldsymbol{x}_i; \boldsymbol{\theta})$. After a softmax function, the probability vector of prediction is $\boldsymbol{f}(\boldsymbol{x}_i; \boldsymbol{\theta})$, i.e.
$\boldsymbol{f}(\boldsymbol{x}_i; \boldsymbol{\theta}) = \text{softmax} (\boldsymbol{h}(\boldsymbol{x}_i; \boldsymbol{\theta}))$.
The objective function is an empirical risk, such as the cross entropy loss, as follows:
\begin{equation}
\begin{aligned}
L(\boldsymbol{\theta}) &= - \frac{1}{n}\sum_{i=1}^n \boldsymbol{y}_i^\top \log \boldsymbol{f}(\boldsymbol{x}_i; \boldsymbol{\theta}).
\end{aligned}
\label{equation: normal_obj}
\end{equation}

In a clean training dataset, the parameters $\boldsymbol{\theta}$ of a DNN is optimized by minimizing \eqref{equation: normal_obj} using gradient descent method. The derivative over the DNN output $\boldsymbol{h}$ is
\begin{equation}
\frac{\partial L}{\partial \boldsymbol{h}} = - \sum_{i=1}^n \big( \boldsymbol{y}_i - \boldsymbol{f}(\boldsymbol{x}_i; \boldsymbol{\theta}) \big).
\label{equation: gradient_over_output}
\end{equation}
With the chain rule, we can update the parameters as
\begin{equation}
\begin{aligned}
\boldsymbol{\theta}_{t+1} = \boldsymbol{\theta}_t - \epsilon \nabla_{\boldsymbol{\theta}_{t}} L(\boldsymbol{\theta}_t) = \boldsymbol{\theta}_t - \epsilon \big( \frac{\partial L}{\partial \boldsymbol{h}} \big)^\top \frac{\partial \boldsymbol{h}}{\partial \boldsymbol{\theta}_t}.
\end{aligned}
\end{equation}

In this work, we consider a classification problem with noisy labels. Let $\tilde{\boldsymbol{y}}_i$ be the noisy label for input $\boldsymbol{x}_i$, and we are only given these noisy labels $\{ \tilde{\boldsymbol{y}}_1, \tilde{\boldsymbol{y}}_2, \cdots, \tilde{\boldsymbol{y}}_n \}$. The connection between clean labels and noisy labels is based on the transition probability,
\begin{equation}
p(\boldsymbol{\tilde{y}} | \boldsymbol{x}) = \sum_{\boldsymbol{y}} p(\boldsymbol{\tilde{y}} | \boldsymbol{y}, \boldsymbol{x}) p(\boldsymbol{y} | \boldsymbol{x}).
\end{equation}
Assuming noise is conditionally independent of inputs \cite{patrini2017making}, given the true labels so that
\begin{equation}
p(\boldsymbol{\tilde{y}} = \boldsymbol{e}^k | \boldsymbol{y} = \boldsymbol{e}^j, \boldsymbol{x}) = p(\boldsymbol{\tilde{y}} = \boldsymbol{e}^k | \boldsymbol{y} = \boldsymbol{e}^j) = T_{jk}.    
\end{equation}
In general, this noise is defined to be class dependent. Noise is uniform with noise rate $\eta$, if
\begin{equation}
\begin{aligned}
    &T_{jk} = 1 - \eta, \text{if } j=k \\
    &T_{jk}=\frac{\eta}{c-1}, \text{if } j \neq k.
\end{aligned}
\label{equation: uniform_noise}
\end{equation}

%-------------------------------------------------------------------------
\subsection{Reflection Loss}

As discussion in previous studies \cite{arpit2017closer,ma2018dimensionality}, there are two stages in training procedure, including a early stage of simple pattern learning and a late stage of label memorization. 
Our work aims at motivating DNNs to take into consideration not only noisy labels, but also simple patterns learned before, which mitigates overfitting to noisy labels. Specifically, we utilize both the original labels $\tilde{\boldsymbol{y}}_i$ and the predictions $\boldsymbol{f}(\boldsymbol{x}_i; \boldsymbol{\theta}_{t-1})$ in the previous epoch as the training target together. The resulting loss (\textit{e.g.,} cross entropy loss) becomes
\begin{equation}
\begin{aligned}
&L(\boldsymbol{\theta}_{t}; \boldsymbol{\theta}_{t-1}) \\
=&- \frac{1}{n} \sum_{i=1}^n \big(\beta \boldsymbol{\tilde{y}}_i + (1 - \beta) \boldsymbol{f}(\boldsymbol{x}_i; \boldsymbol{\theta}_{t-1}) \big)^\top \log \boldsymbol{f}(\boldsymbol{x}_i; \boldsymbol{\theta}_t) \\
=& \beta L_o + (1 - \beta) L_r,
\end{aligned}
\label{equation: our_loss}
\end{equation}
where the former $L_o$ is the original loss in clean setting and the latter $L_r$ provides resistance to label noise. The weighted coefficient, $\beta$, is the ``confidence'' of annotation quality. If we are more confident in the label quality (\textit{i.e.,} larger $\beta$), DNN relies more on original raw labels. Otherwise, DNN refines more from its last predictions, corresponding to its learned patterns.

Since we utilize a convex combination of labels in the dataset and predictions in the previous epoch, our reflection loss belongs to  pseudo-label method~\cite{reed2014training,ma2018dimensionality,tanaka2018joint}. After separating our loss in \eqref{equation: our_loss}, the second part can be regarded as a special consistency loss, in which we replace mean squared error (MSE) with cross entropy (CE) and utilize predictions only in the previous epoch instead of ensembling historical predictions.

%-------------------------------------------------------------------------
\subsection{Gradient Analysis}

\begin{figure}[!t]
	\begin{center}
		\includegraphics[width=0.8\linewidth]{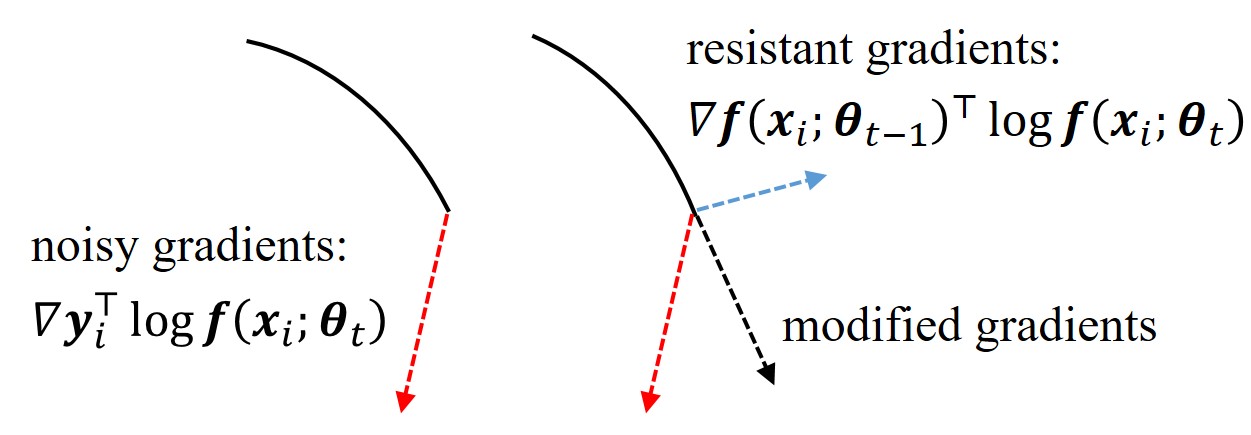}
	\end{center}
	\caption{The illustration of gradient correction. Left: gradients from noisy samples may lead to overfit; Right: noisy gradients is modified by correction gradients, which is less prone to overfit to noisy labels.}
	\label{fig:diagram_for_gradients}
\end{figure}

In this part, we explore the mechanism behind our proposed loss and discuss its merits. In conventional gradient descent algorithm, gradients calculated from noisy samples may lead to overfitting \cite{li2018learning} as shown in Fig. \ref{fig:diagram_for_gradients}. Notably, our method provides a resistance to these noisy gradients, which is apparent after derivation of the gradient over the DNN outputs $\boldsymbol{h}$ for a given sample $\boldsymbol{x}$:
\begin{equation}
\begin{aligned}
\frac{\partial L}{\partial \boldsymbol{h}}
= &-(1 - \beta) \sum_i \big( \boldsymbol{f}(\boldsymbol{x}_i; \boldsymbol{\theta}_{t - 1}) - \boldsymbol{f}(\boldsymbol{x}_i; \boldsymbol{\theta}_t) \big)\\
& \qquad - \beta \sum_i \big( \tilde{\boldsymbol{y}}_i -  \boldsymbol{f}(\boldsymbol{x}_i; \boldsymbol{\theta}_t) \big).
\end{aligned}
\label{equation: gradient_for_our_method}
\end{equation}

Other correction methods, such as bootstrap method \cite{reed2014training}, have similar decomposition and have different resistance mechanism from ours.
Given a sample $\boldsymbol{x}$, for bootstrap-soft method, we derivate the gradients of the pesudo-label term over the output $h_i$ (before softmax) of DNN for class $l$: 
\begin{equation}
    \frac{\partial L}{\partial h_l} = f_l (\sum_j f_j \log f_j - \log f_l),
\label{eqn:gradient_soft}
\end{equation}
where $f_l$ is the output probability (after softmax) of class $l$, \textit{i.e.} $f_l^t = f_l(\boldsymbol{x}; \boldsymbol{\theta}_t)$. 

For boostrap-hard, the result is
\begin{equation}
    \frac{\partial L}{\partial h_l} = f_l^t - \mathbb{I}(l = \arg\max_j f_j^t).
\end{equation}
Assuming that we have a sample with prediction of class $k$, that is $f_k \geq f_l$, the gradients for different class satisfying
\begin{equation}
    \frac{\partial L}{\partial h_k} \geq \frac{\partial L}{\partial h_l}, l=1, 2, \cdots, c
\end{equation}
which means the pesudo-label term push the predictions more confident.

For our reflection loss, the gradients is
\begin{equation}
    \frac{\partial L}{\partial h_l} = f_l^t - f_l^{t-1}.
\label{eqn:gradient_TCR}
\end{equation}
Obviously, Equation \eqref{eqn:gradient_TCR} achieves satisfying performance without pushing the prediction more confidence, which makes it is suitable for open-set noise \cite{wang2018iterative}.

%-------------------------------------------------------------------------
\subsection{Squeeze Technique}

Note that the resistance in reflection loss will degrade as the learning rate $lr$ decays, which is also called the \textit{Resistance Degradation Problem}. For a given sample $\boldsymbol{x}_i$, the resistance in \eqref{equation: gradient_for_our_method} is $\boldsymbol{f}(\boldsymbol{x}_i; \boldsymbol{\theta}_{t - 1}) - \boldsymbol{f}(\boldsymbol{x}_i; \boldsymbol{\theta}_t)$, which is related to the step size. Even with same gradients to descend, smaller step size results in smaller difference $\boldsymbol{f}(\boldsymbol{x}_i; \boldsymbol{\theta}_{t - 1}) - \boldsymbol{f}(\boldsymbol{x}_i; \boldsymbol{\theta}_t)$, which degrades the resistance in next iteration. As a result, the resistance is no longer enough to overcome overfitting.

Thus, we propose a technique called ``Squeeze'' to keep enough resistance after learning rate decays. Since the DNN has learned some simple pattern, we could make DNN be more confident in its predictions. Assuming the probability for a sample is $\boldsymbol{p}$, the squeeze function is
\begin{equation}
\text{Squeeze}(\boldsymbol{p}) = \frac{\boldsymbol{p}^\gamma}{\boldsymbol{1}^\top \boldsymbol{p}^\gamma},
\end{equation}
where $(\cdot)^\gamma$ is the element-wise power to $\gamma \geq 1$. The squeeze function becomes one-hot function as $\gamma \rightarrow \infty$. With this function, largest value in prediction vector becomes a bit larger, so as to provide larger gradient for the most probable class and smaller ones for potential noisy samples.

Compared with other techniques to improve DNN confidence (\textit{e.g.}, minimum entropy regularization in \cite{tanaka2018joint}), we only modify labels and keeps others unchanged. Besides, entropy regularization is severely affected by the number of classes, which makes hyper-parameter settings hard to extend, e.g. from CIFAR-10 to CIFAR-100, while our method could remain the same hyper-parameters. 

%-------------------------------------------------------------------------
\subsection{The Full Algorithm of TCR}

\begin{figure}[!t]
 \removelatexerror
  \begin{algorithm}[H]
   \caption{Temporal Calibrated Regularization}
   Randomly initialize $\boldsymbol{\theta}$\;
   $Z = [\boldsymbol{z}_1, \cdots, \boldsymbol{z}_n] \leftarrow \boldsymbol{0}_{[c \times n]}$ \;
   \For{$t $ in $[1, \text{num\_epochs}]$}
   {
      \For{each mini-batch $B$}{
         $\boldsymbol{z}_i^{(t)} \leftarrow \boldsymbol{f}(\boldsymbol{x}_i, \boldsymbol{\theta})$ \;
         \eIf{$t \geq 1$}
         {
            $\boldsymbol{y}_i^* \leftarrow \beta \boldsymbol{y}_i + (1 - \beta) \boldsymbol{z}_i$
         }{
            $\boldsymbol{y}_i^* \leftarrow \boldsymbol{y}_i$
         }
         $L \leftarrow \sum_{i \in B} {\boldsymbol{y}_i^*}^\top \log \boldsymbol{f}(\boldsymbol{x}_i, \boldsymbol{\theta}) $ \;
         update $\boldsymbol{\theta} \leftarrow \boldsymbol{\theta}_{t} - \epsilon \nabla_{\boldsymbol{\theta}} L$ \;
         $\boldsymbol{z}_i \leftarrow \boldsymbol{z}_i^{(t)} $ \;
         \If{$t \geq T_{s}$}
         {
            $\boldsymbol{z}_i \leftarrow \text{Squeeze}(\boldsymbol{z}_i)$
         }
      }
   }
   \Return{$\boldsymbol{\theta}$}
  \end{algorithm}
\end{figure}

The detailed procedure is outlined in Algorithm 1. Before updating the parameters, we obtain the current predictions of the DNN (Line 5). After the update, we record these prediction in the storage $z$. The pseudo-labels for loss is the convex combination of original labels and predictions recorded in the storage, \textit{i.e.}, the prediction in the previous epoch (Line 6-7). Since there is no last predictions for the first epoch, we use the original labels in first epoch (Line 8-9). To improve the DNN confidence over its predictions and compensate the degraded gradients after learning rate decays, we utilize the ``Squeeze'' function. TCR owns the following good properties:

\noindent\textbf{Easy Implementation} Note that, TCR is a general extension and easy to implement requiring few modifications. All we need is to simply add two parts, a Temporal Delay and a Squeeze technique, in conventional models as shown in Fig. \ref{fig:diagram_of_our_method}. The former stores predictions in current epoch and provides its collections for DNN in next epoch.

\noindent\textbf{Less Time and Computation Cost} Compared with the prior work \cite{ren2018learning,jenni2018deep,li2018learning} in gradient correction under label noise, our method is just a label correction method based on prediction difference between adjacent epochs, and has less overhead. The only storage overhead in our method is the memory for predictions of DNN in the previous epoch, which is proportional to $O(c \cdot n)$. The time overhead in our method is almost negligible.

%-------------------------------------------------------------------------

\section{Experiments}

In this section, we firstly conduct a series of comparative experiments to empirically understanding our proposed method. We then evaluate the robustness of our proposed model to noisy labels with comprehensive experiments on different noise types and different noise ratio.

%-------------------------------------------------------------------------
\subsection{Understanding the Learning Process}

\begin{figure}[t]
    \centering
  \subfloat[a\label{1a}]{%
       \includegraphics[width=0.48\linewidth]{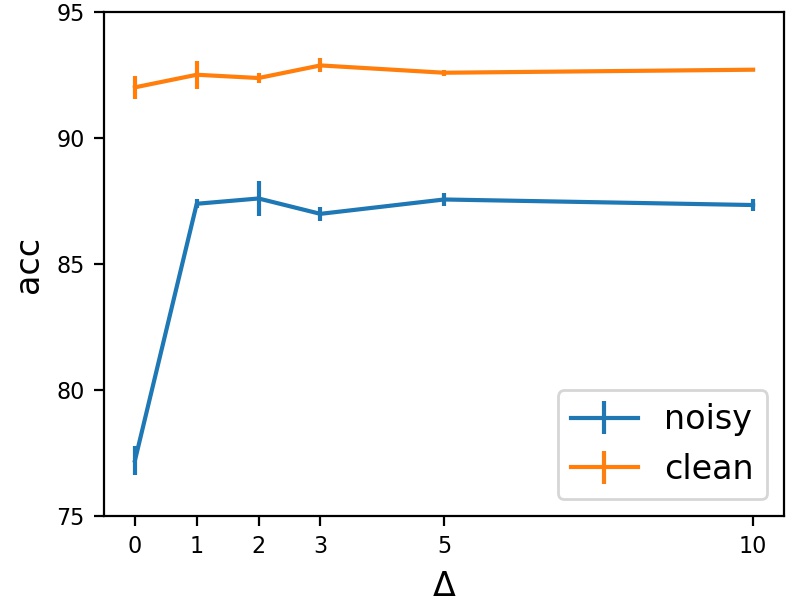}}
    \hfill
  \subfloat[b\label{1b}]{%
        \includegraphics[width=0.48\linewidth]{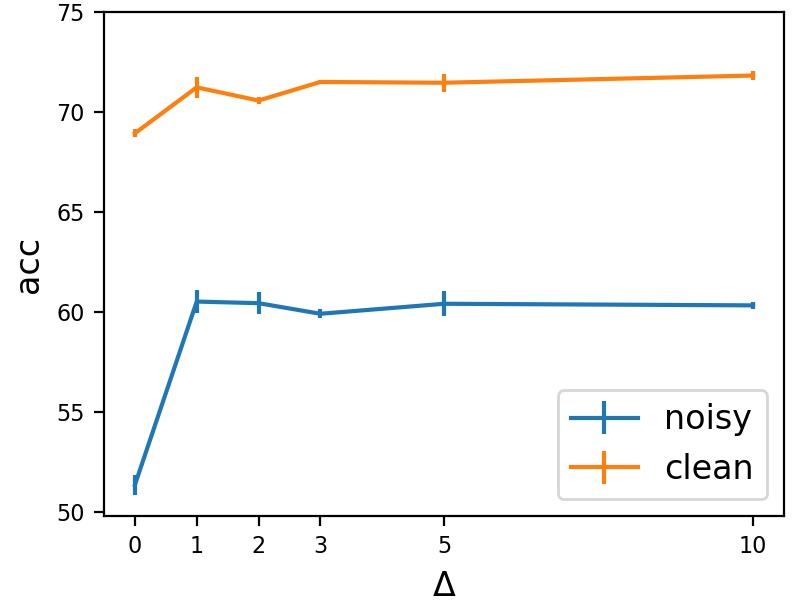}}
  \caption{Accuracy(\%) on clean and noisy (40\% uniform noise) CIFAR-10 and CIFAR-100 with variable time difference $\Delta$. As long as we use the predictions before, we achieve benefits. And the time difference has negligible effects.}
  \label{fig: ablation_delay} 
\end{figure}

Recall that we use the predictions in the previous epoch, we test whether to obtain more from predictions of earlier times. We test the results of ResNet-32 in clean and noisy (40\% uniform noise) CIFAR-10 and CIFAR-100 with varying time difference $\Delta = \{0, 1, 2, 3, 5, 10\}$, \textit{i.e.}, in the current epoch $t$, we utilize the prediction in the epoch $t - \Delta$. As shown in Fig. \ref{fig: ablation_delay}, there is negligible difference between different $\Delta$, except the case $\Delta=0$, which actually uses current predictions. The results indicate that the determining factor is to let DNN consider its previous predictions, and how long does not matter. From a perspective of implementation, the space complexity is $O(n \cdot \Delta)$, large $\Delta$ takes more challenge in storage. In a word, $\Delta=1$ has the satisfying performance and lowest space complexity.

%-------------------------------------------------------------------------

\begin{figure}[t]
    \centering
  \subfloat[40\% uniform noise]{%
       \includegraphics[width=0.48\linewidth]{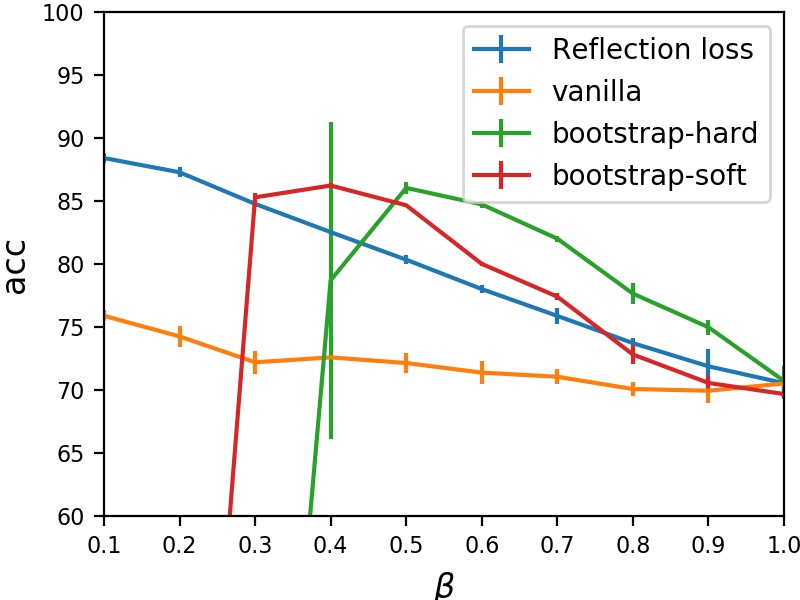}}
    \hfill
  \subfloat[40\% asymmetric noise]{%
        \includegraphics[width=0.48\linewidth]{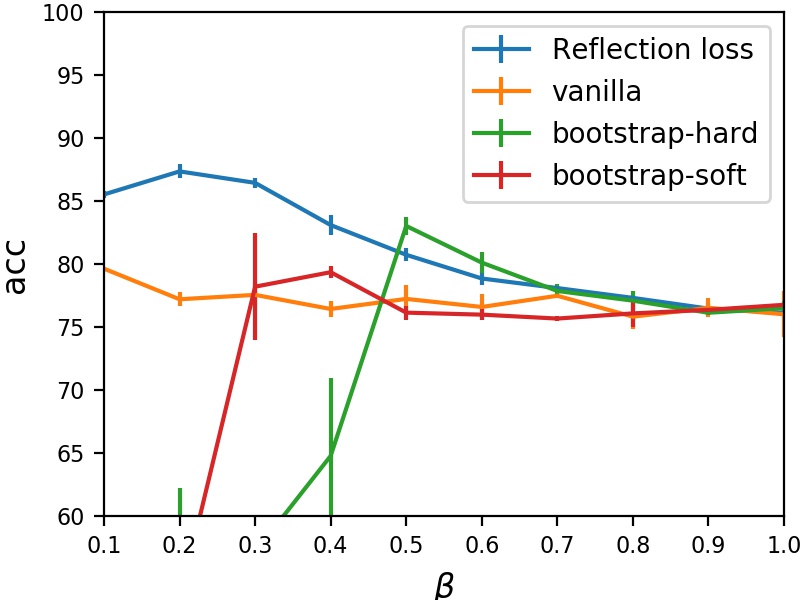}}
  \caption{Accuracy of different label correction methods on noisy CIFAR-10 with varying hyper-parameter $\beta$. The vanilla method just reduce the learning rate and has little effect. The bootstrap-hard and bootstrap-soft are sufficient but fragile as $\beta$ is small. Our reflection loss is stable in different choices of $\beta$.}
  \label{fig: ablation_beta} 
\end{figure}

\subsection{Comparison to Label Correction Methods}
\noindent\textbf{Experimental setup:} We utilize the PreAct-ResNet18. The networks are trained on 40\% uniform noise. All experiments use the mini-batch of size 128 and networks were trained using SGD with momentum 0.9, weight decay $10^{-4}$ and an initial learning rate of 0.1. The learning rate is divided by 10 after epochs 80, 120 (150 epochs in total). As we have discussed the difference to correction methods, we now empirically compare their performance to show the advantage of TCR in CIFAR-10 with 40\% uniform noise or 40\% asymmetric noise. Since different methods may have different choices of hyper-parameter $\beta$, we test the performance in $\beta = \{1.0, 0.9, 0.8, 0.7, 0.6, 0.5, 0.4, 0.3, 0.2, 0.1\}$.

\noindent\textbf{Results:}
Fig. \ref{fig: ablation_beta} demonstrates the the influence of different correction methods. We can observe that the vanilla method which uses the current predictions has a slightly better results. The correction methods like bootstrap-soft and bootstrap-hard have satisfying performance only in specific $\beta$, since they have to keep balance between enough resistance and pattern learning in the beginning. For our method, it can learn pattern and keep enough resistance in smaller $\beta$ as discussed before.

\begin{figure*}[t]
    \centering
  \subfloat[an ``airplane''(row 0) with a label ``automobile''(row 1)]{%
      \includegraphics[width=0.4\linewidth]{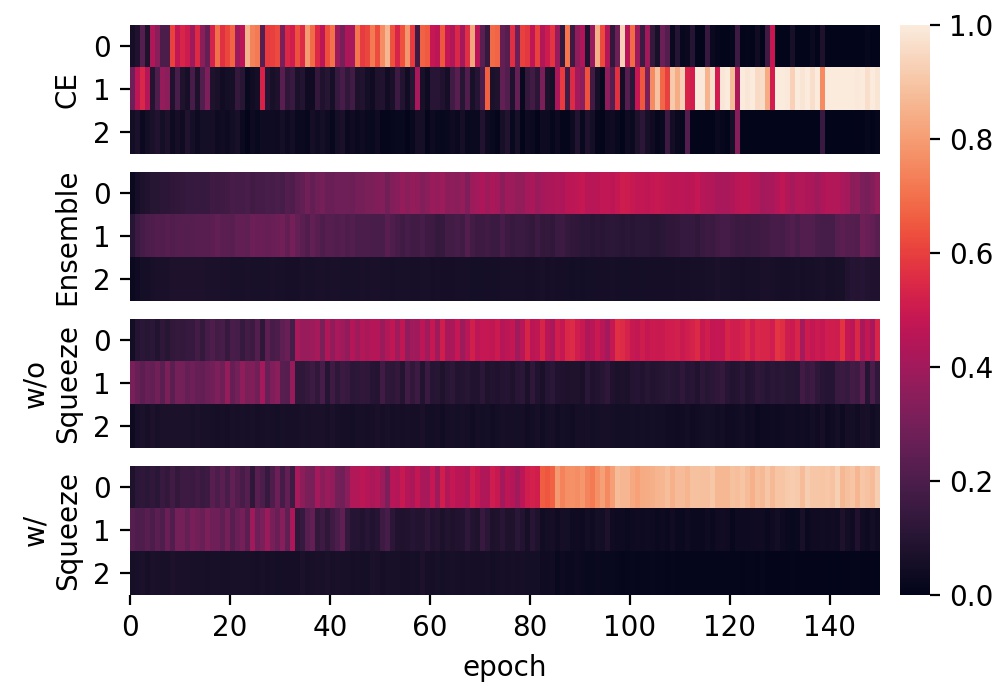}}
    % \hfill
  \subfloat[an ``airplane''(row 0) with a correct label]{%
        \includegraphics[width=0.4\linewidth]{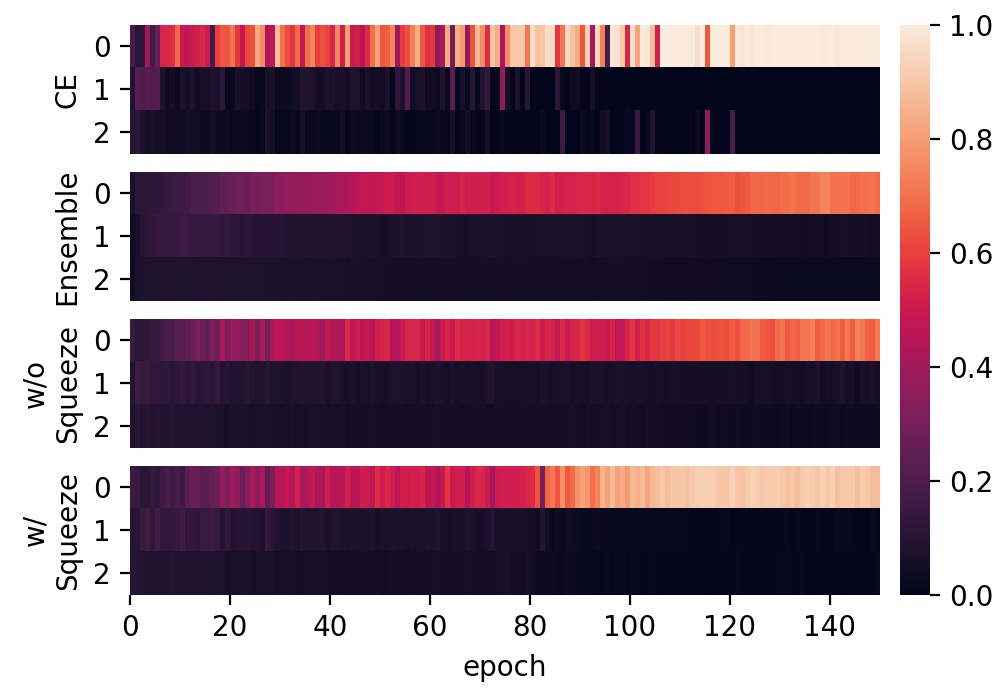}}
  \caption{We visualize the predicted probability of DNN during training for different losses. From the top to bottom, they cross entropy loss, reflection loss with EWA, the original reflection loss and our method. The conventional approach with CE learns simple pattern first but memorizes the noisy label at last. while our methods learn simple pattern and keeps it until the end.}
  \label{fig: outputs_visualization} 
\end{figure*}

%-------------------------------------------------------------------------
\subsection{Comparison to Temporal Ensemble}

Although our reflection loss can be regarded as a special case of the consistency loss in Temporal Ensemble \cite{laine2016temporal}, in which we replace the Mean Square Error with Cross Entropy and only utilize predictions in the previous epoch. We empirically demonstrate that the latter is a significant setting.

\begin{figure}[t]
    \centering
  \subfloat[CIFAR-10]{
       \includegraphics[width=0.48\linewidth]{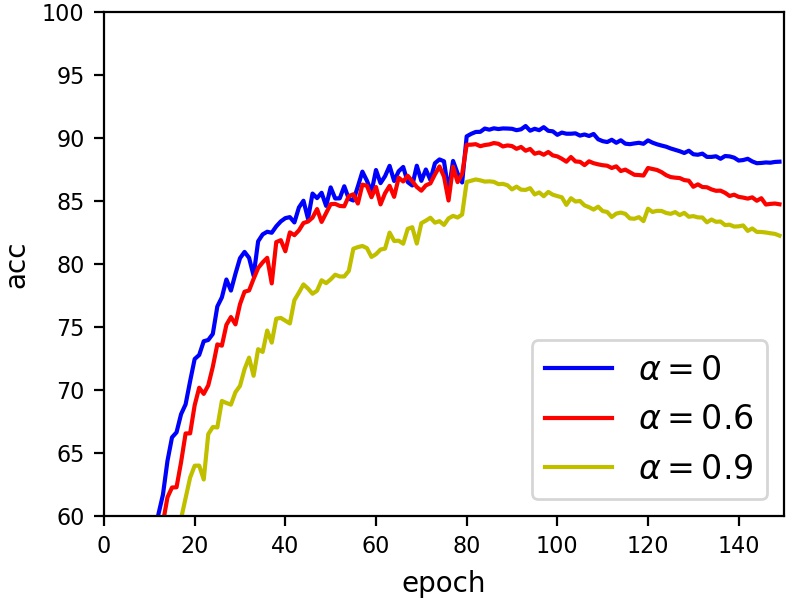}}
    \hfill
  \subfloat[CIFAR-100]{
        \includegraphics[width=0.48\linewidth]{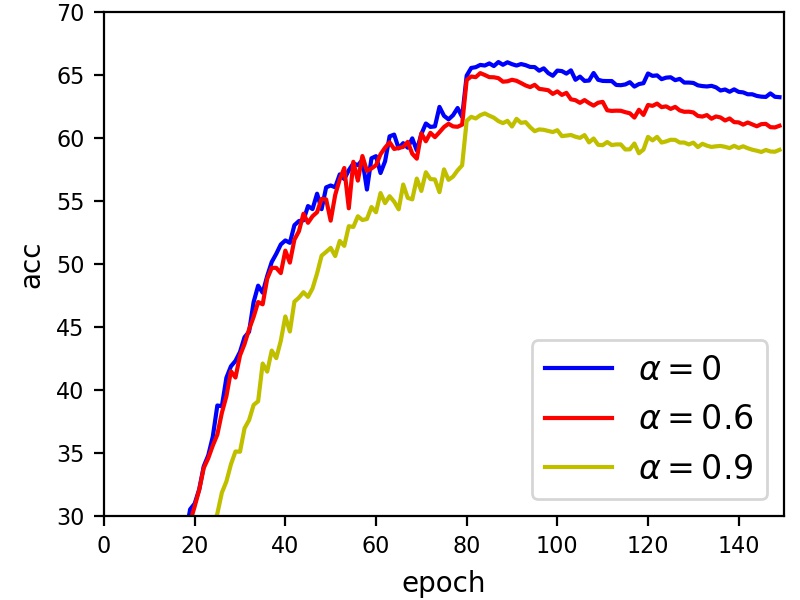}}
  \caption{Accuracy in clean test set during training on CIFAR-10 with 40\% uniform noise. If we set $\alpha$ a large value (0.9), it hampers learning simple pattern in early training since the poor predictions early influence longer. If we set $\alpha=0.6$, it has learned simple pattern in early stage, but start overfitting severely after learning rate decays.}
  \label{fig: ensemble} 
\end{figure}

The original Temporal Ensemble utilizes the Exponentially Weighted Average (EWA) to average predictions as\footnote{Here, we denote $\boldsymbol{f}(\boldsymbol{x}_i, \boldsymbol{\theta}_t)$ as $\boldsymbol{f}^{(t)}_i$ for simplicity.}
\begin{equation}
(1 - \alpha)\boldsymbol{f}^{(t)}_i + (1 - \alpha) \alpha \boldsymbol{f}^{(t-1)}_i (1 - \alpha) \alpha^2 \boldsymbol{f}^{(t-2)}_i + \cdots,
\end{equation}
while we find it harms the learning from two aspects: 1) slowing down the learning process in early stage; 2) making DNN easier to overfitting. To validate this, we utilize the same setting in Section 4.2 except for replacing the predictions in the previous epoch with the ensemble predictions and set $\alpha=\{0, 0.6, 0.9\}$ respectively. The accuracy in test set is plotted in Fig. \ref{fig: ensemble}.

If we set $\alpha$ a large value (0.9), it hampers learning simple pattern in early training since the poor predictions early influence longer. If we set $\alpha=0.6$, it has learned simple pattern in early stage, but start overfitting severely after learning rate decays. As shown in Fig. \ref{fig: outputs_visualization}, the predicted probabilities for model with $\alpha=0.6$ are more smoothing, while ours are fluctuate, which have larger difference in temporal dimension. According to \eqref{equation: gradient_for_our_method}, the resistant gradient is related to the difference in temporal dimension. Thus, $\alpha=0$ adds more noise in gradients comparing to $\alpha=0.6$, which improves its robustness.

%-------------------------------------------------------------------------
\subsection{The Effect of Squeeze Technique}

\begin{figure}[t]
    \centering
  \subfloat[CIFAR-10]{%
       \includegraphics[width=0.48\linewidth]{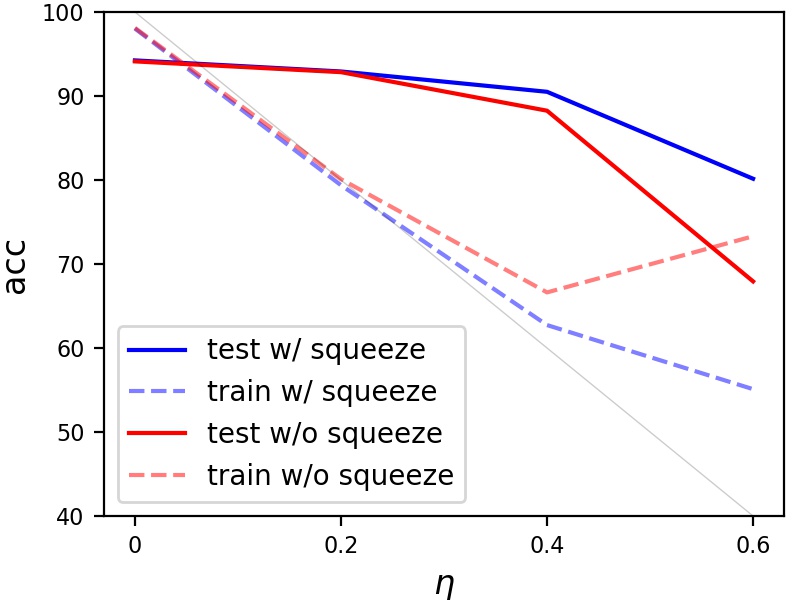}}
    \hfill
  \subfloat[CIFAR-100]{%
        \includegraphics[width=0.48\linewidth]{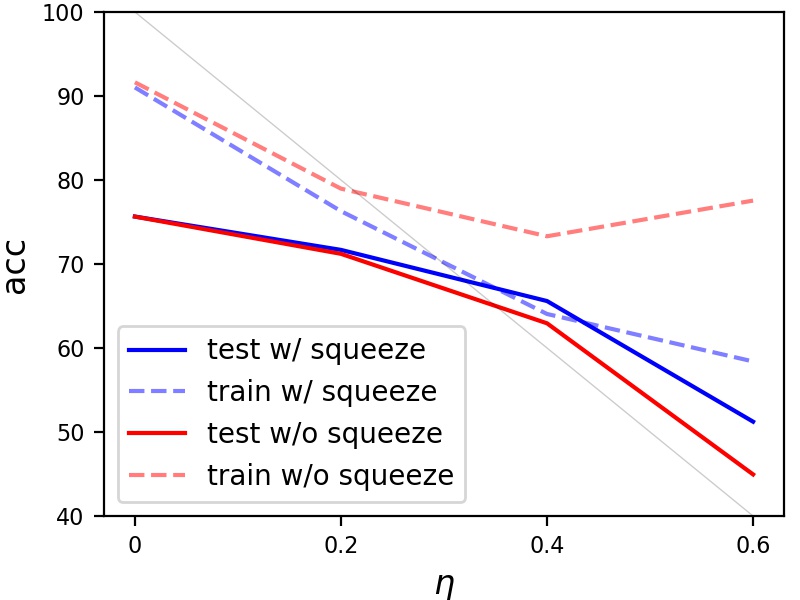}}
  \caption{Accuracy(\%) on training set and test set of CIFAR-10 with varying ratio of uniform noise. As the noise ratio increase, DNN without ``Squeeze'' overfits to noisy labels severely (red dashed line), and the test accuracy decreases (red solid line). ``Squeeze'' impedes this problem effectively as shown in blue line.}
  \label{fig: squeeze_effect} 
\end{figure}

To validate the significance of the squeeze function in our method, we conduct a series of experiments. First, we visualize the predicted probability for samples with a correct or incorrect label in Fig. \ref{fig: outputs_visualization}. For a DNN only with our reflection loss (no squeeze), its prediction confidence is not high enough whether the label is correct or not. As introducing the squeeze function after the first learning rate decay, its prediction confidence becomes higher. Next, we plot the accuracy in noisy train set and clean test set during training, as shown in Fig. \ref{fig: squeeze_effect}. In severe uniform noise, the reflection loss encounters memorizing noisy samples (accuracy in noisy train set is larger than the ratio of correct samples), and the squeeze technique overcomes this problem significantly.

%-------------------------------------------------------------------------
\subsection{Robustness against Noisy Labels}

Finally, we evaluate the robustness of our method against noisy labels under different conditions (various noise types and ratios, architectures, etc.) to demonstrate the advantageous performance of our method.

\begin{table*}[t]
	\centering
	% \small
	    \caption{Average accuracy (\%) in clean test set (5 runs) with varying noise ratios in uniform noise and asymmetric noise. The best results are highlighted in \textbf{bold}. Since Forward Loss requires the ground truth confusion matrix, we also highlight the second one in \textbf{bold} if the result of Forward Loss is the highest one.}
		\begin{tabular}{llcccccccc}
			\hline
			\multirow{2}{*}{Datasets} & \multirow{2}{*}{Method} & \multirow{2}{*}{clean} & \multicolumn{3}{c}{Symmetric Noise} & \multicolumn{4}{c}{Asymmetric Noise} \\ \cline{4-10} 
			&  &  & 0.2 & 0.4 & 0.6 & 0.1 & 0.2 & 0.3 & 0.4 \\ \hline
			\multirow{7}{*}{CIFAR-10} & CE & 94.31 & 80.13 & 60.75 & 39.70 & 90.84 & 86.19 & 81.98 & 76.83 \\
			& Forward & 94.31 & 86.37 & 76.52 & 60.49 & 92.55 & 90.06 & 88.67 & 86.06 \\
			& GCCE($q=0.7$) & 93.35 & 91.51 & 87.18 & 68.04 & 92.04 & 89.75 & 82.78 & 76.32 \\
			& mixup($\alpha=8$) & \textbf{94.46} & 92.17 & 88.55 & 80.14 & 93.77 & 92.56 & 90.94 & 86.61 \\
			& Joint & 91.49 & 90.34 & 89.41 & 79.68 & 92.45 & 92.28 & 91.32 & 90.30 \\
			& Ours w/o Squeeze & 94.16 & 92.84 & 88.25 & 67.93 & 93.60 & 93.09 & 92.12 & 89.11 \\
			& Ours & 94.24 & \textbf{92.92} & \textbf{90.50} & \textbf{80.15} & \textbf{93.90} & \textbf{93.16} & \textbf{92.52} & \textbf{90.53} \\ \hline
			\multirow{7}{*}{CIFAR-100} & CE & 74.30 & 59.38 & 43.71 & 24.53 & 68.35 & 61.25 & 53.91 & 44.42 \\
			& Forward & 74.50 & 63.94 & 51.66 & 37.66 & 73.45 & 73.02 & \textbf{72.18} & \textbf{71.92} \\
			& GCCE($q=0.7$) & 69.76 & 67.32 & 64.08 & 50.70 & 68.58 & 66.75 & 64.22 & 51.84 \\
			& mixup($\alpha=8$) & 74.33 & 68.68 & 58.56 & 42.78 & 71.44 & 67.45 & 61.12 & 50.06 \\
			& Ours w/o Squeeze & 75.64 & 71.21 & 62.94 & 44.95 & 74.52 & 72.65 & 67.73 & 56.06 \\
			& Ours & \textbf{75.65} & \textbf{71.68} & \textbf{65.59} & \textbf{51.23} & \textbf{74.95} & \textbf{73.81} & \textbf{71.19} & \textbf{63.15} \\ \hline
		\end{tabular}
\label{table: types_and_ratios}
\end{table*}

\noindent\textbf{Experimental setup:}
Experiments were conducted on widely used datasets: CIFAR-10 and CIFAR-100 with different architectures like ResNet34, ResNet44, PreAct ResNet-18 and Wide ResNet (WRN). All Networks are trained using SGD with momentum 0.9, weight decay $10^{-4}$ and an initial learning rate of 0.1. The learning rate is divided by 10 after epochs 80, 120 (150 epochs in total) except for Wide ResNet. We train WRN with mini-batch size 100, and divide its learning rate after epochs 80, 100 (120 epochs in total), which is same as the description in \cite{ren2018learning}. Simple data augmentations (width/height shift and horizontal flip) are applied. For our method, TCR, we all set $\beta=0.1, \gamma=1.1$ in each experiment of this section.

%-------------------------------------------------------------------------

\noindent\textbf{Baselines:} We compare our method with state-of-the-art baselines for noisy label learning. The baselines include: 1) Cross Entropy Loss: the conventional approach of training with cross-entropy loss; 2) Forward Loss \cite{patrini2017making}: a robust loss through multiplying the prediction of DNN by a estimated label transition probability matrix; 3) Generalized Cross Entropy Loss (GCCE) \cite{zhang2018generalized}: a noise-robust loss with a hyper-parameter $q$, which can be seen as a generalization of cross-entropy loss ($q=0$) and mean absolute error ($q=1$). Here, $q=0.7$ is used as suggested in \cite{zhang2018generalized}; 4) mixup \cite{zhang2017mixup}: a special data augmentation using a linear combination of images and labels. We tested $\alpha = \{1, 4, 8, 32\}$, and set $\alpha=8$ which has the best results; 5) Joint \cite{tanaka2018joint}: a joint optimization framework of learning DNN parameters and estimating true labels. Here, we use the first part of Joint since other methods are only trained once. The parameters are set according to the original paper, $lr=0.08, \alpha=1.2, \beta=0.8 $ for symmetric noise, $lr=0.03, \alpha=0.8, \beta=0.4$ for asymmetric noise.

\noindent\textbf{Robustness to Partially Corrupted Labels:} Asymmetric noise and uniform (symmetric) noise are typical settings of partially corrupted labels. Both noises are generated according to \cite{patrini2017making}. All experiments are conducted with PreAct ResNet-18. We report the average accuracy over 5 repetitions of the experiments in Table \ref{table: types_and_ratios}. For noisy CIFAR-10, our proposed method outperforms the baselines significantly. Actually, the ``reflection'' loss already has a comparative performance to mixup. However, our method is little worse in clean training data, which we conjecture is because they resist to memorize all training samples and the training accuracy cannot be close to 100\% in only 150 epochs. For noisy CIFAR-100, our method has the best performance overall. Note that Forward sometimes also delivers a relatively good performance, as we directly provide it with the ground truth noise matrix which is not often available in real-world settings.  
In conclusion, our method has advantageous performance and generality for different noise types and ratios.

%-------------------------------------------------------------------------
\noindent\textbf{Extension to Different Architectures:} For more comprehensive comparison with other methods proposed recently and validation of the generality of our method, we conduct experiments with different architectures, including ResNet-44 \cite{he2016deep}, ResNet-34 \cite{he2016deep}, PreAct ResNet-18 \cite{he2016identity} and WRN \cite{zagoruyko2016wide}. We also list the reported accuracy from other state-of-art methods, such as D2L \cite{ma2018dimensionality}, GCCE \cite{zhang2018generalized}, Bilevel \cite{jenni2018deep}, mixup \cite{zhang2017mixup} (our reproduced implementation) and a MAML method \cite{ren2018learning}. The results are shown in Table \ref{table: architectures}. Our method has consistently
robustness with all architectures. In every architectures,  ours outperform its corresponding competitors. For different architectures, we all set the hyper-parameters $\beta=0.1, \gamma=1.1$, which indicates our method is easy to extend.

\begin{table}[t]
    \centering
    % \small
    \caption{Accuracy (\%) in clean test set of different architectures with 40\% uniform noise. We also list the reported accuracy from other state-of-art methods.}
	\begin{tabular}{llcc}
		\hline
		Method & Architecture & CIFAR-10 & CIFAR-100 \\ \hline
		\multirow{4}{*}{CE} & ResNet-44 & 66.17 & 45.71 \\
		& ResNet-34 & 65.45 & 45.30 \\
		& PreAct ResNet-18 & 60.75 & 43.71 \\
		& WRN & 78.28 & 50.40 \\ \hline
		D2L & ResNet-44 & - & 52.01 \\
		GCCE & ResNet-34 & 87.13 & 61.77 \\
		Bilevel & PreAct ResNet-18 & 87.0 & 59.8 \\
		mixup & PreAct ResNet-18 & 88.55 & 58.56 \\
		MAML & WRN & 86.92 & 61.34 \\ \hline
		\multirow{4}{*}{Ours} & ResNet-44 & 89.60 & 58.13 \\
		& ResNet-34 & 90.72 & 64.58 \\
		& PreAct ResNet-18 & 90.50 & 65.59 \\
		& WRN & 91.43 & 67.35 \\ \hline
	\end{tabular}
\label{table: architectures}
\end{table}

%-------------------------------------------------------------------------
\noindent\textbf{Performance on Open-set Noise:}
Open-set noise \cite{wang2018iterative} is another common label noise in real data, in which a noisy sample possesses a true class that is not contained within the set of known classes. When we collect data through search engines, it is impossible to ensure the noisy data only from the known classes we have interests in. We use the open-set noisy dataset, CIFAR-10 + CIFAR-100 as described in \cite{wang2018iterative}. The result is shown in Table \ref{table: open_set} and our method outperforms its competitors. In open-set noise, it is not a good choice to push predictions to be more confident, as some images do not belong to any class. This may be the reason that bootstrap-hard (boots in Table \ref{table: open_set}) has worsen performance than ours. 

\begin{table}[t]
	\centering
	\caption{Accuracy (\%) in clean test set with 40\% open-set noise.}
	\begin{tabular}{lccccc}
		\hline
		noise type & CE & GCCE & boots & mixup & ours \\ \hline
		CIFAR100 & 86.73  & 87.52 & 88.26 & 91.97 & 92.29 \\ \hline
	\end{tabular}
\label{table: open_set}
\end{table}

\begin{table}[t]
	\centering
	% \small
	\caption{Test accuracy (\%) on the Clothing1M dataset}
	\begin{tabular}{lccccc}
		\hline
		Method & CE & Forward & Bilevel & Joint & Ours \\ \hline
		Acc & 68.94 & 69.84 & 69.9 & 72.16 & 72.54 \\ \hline
	\end{tabular}
\end{table}

%-------------------------------------------------------------------------
\noindent\textbf{Experiments on Real-World Data with Noisy Labels:}
Lastly, we test our method on the real-world large-scale noisy dataset Clothing1M \cite{xiao2015learning}, which consists of fashion images belonging to 14 classes: T-shirt, Shirt, Knitwear, Chiffon, Sweater, Hoodie, Windbreaker, Jacket, Down Coat, Suit, Shawl, Dress, Vest, and Underwear. The labels are generated by the surrounding text of images and are thus extremely noisy. The overall accuracy of the labels is ${\small\sim61.54\%}$, with some pairs of classes frequently confused with each other (\textit{e.g.}, Knitwear and Sweater), which may contain both symmetric and asymmetric label noise. We used ResNet-50 pre-trained on ImageNet to align experimental condition with \cite{patrini2017making}. For preprocessing, we resize the image to $256 \times 256$, crop the middle $224 \times 224$ as input, and perform normalization. We use a batch size $k = 32$, learning rate $lr=0.0008$, and update using SGD with momentum $0.9$ and weight decay $10^{-3}$.

We utilize 3 epochs to train DNN on noisy data, 1 epoch for normal training, 1 epoch for only reflection loss, and 1 epoch for the full algorithm. We still set $\beta=0.1, \gamma=1.1$. We achieve 72.54\% at last, which is better than 68.94\% from Cross Entropy Loss (quoted from \cite{patrini2017making}), 69.8\% from Forward Loss \cite{patrini2017making}, 69.9\% from Bilevel \cite{jenni2018deep}, and  72.16\% from \cite{tanaka2018joint}. 

%-------------------------------------------------------------------------

\section{Conclusion}

In this paper, we proposed a label-correction method, called Temporal Calibrated Regularization (TCR), to let DNN consider the simple pattern it has learned and labels in dataset simultaneously, so as to overcome the overfitting to noisy labels. Specifically, TCR utilizes the convex combination of original labels and predictions in the previous epoch as pseudo-labels and applies ``Squeeze'' to improve the DNN confidence for its predictions. Experiments on various network architectures and datasets demonstrated that our method significantly outperforms the state-of-the-arts. 

%-------------------------------------------------------------------------

\section{Acknowledgment}

Shu-Tao Xia is supported in part by National Key Research and Development Program of China under Grant 2018YFB1800204, National Natural Science Foundation of China under Grant 61771273,
R\&D Program of Shenzhen under Grant JCYJ20180508152204044, and research fund of PCL Future Regional Network Facilities for Large-scale Experiments and Applications (PCL2018KP001).

\bibliographystyle{IEEEtran}
\bibliography{egbib}

\end{document}